\documentclass[runningheads]{llncs}
\usepackage{comment}
\usepackage{graphicx}
\usepackage{color}
\usepackage{url}
\usepackage{subfigure} 
\usepackage{multibib}
\usepackage[leqno]{amsmath}
\usepackage{examples}
\usepackage{multirow}
\usepackage{xcolor}

\usepackage[subtle]{savetrees}

\iftrue

\newcommand{\EK}[1]{\textcolor{red}{[EK: #1]}}
\newcommand{\VD}[1]{\textcolor{violet}{[VD: #1]}}

\else

\newcommand{\EK}[1]{\textcolor{red}{[EK: \#1]}}
\newcommand{\VD}[1]{\textcolor{violet}{[VD: \#1]}}

\fi

\begin{document}

%
\title{Automated Personalized Feedback Improves Learning Gains in an Intelligent Tutoring System}
\titlerunning{Personalized Feedback Improves Learning Gains in an ITS}
%
\author{Ekaterina Kochmar\inst{1,2} \and Dung Do Vu\inst{1,3} \and Robert Belfer\inst{1} \and Varun Gupta\inst{1} \and \\ Iulian Vlad Serban\inst{1} \and Joelle Pineau\inst{1,4}
}
%
\authorrunning{Kochmar et al.}
%
\institute{
Korbit Technologies Inc.
\and University of Cambridge \and
{\'E}cole de Technologie Sup{\'e}rieure \and
McGill University \& MILA (Quebec Artificial Intelligence Institute) \\}
%
\maketitle
%
\vspace{-1.5em}
\begin{abstract}

We investigate how automated, data-driven, personalized feedback in a large-scale intelligent tutoring system (ITS) improves student learning outcomes.
We propose a machine learning approach to generate personalized feedback, which takes individual needs of students into account.
We utilize state-of-the-art machine learning and natural language processing techniques to provide the students with \textit{personalized hints}, \textit{Wikipedia-based explanations}, and \textit{mathematical hints}.
Our model is used in {\tt Korbit},\footnote{\url{https://www.korbit.ai}} a large-scale dialogue-based ITS with thousands of students launched in 2019, and we demonstrate that the personalized feedback leads to considerable improvement in student learning outcomes and in the subjective evaluation of the feedback.



\end{abstract}

\section{Introduction}

Intelligent Tutoring Systems (ITS)~\cite{anderson1985intelligent,nye2014autotutor}
attempt to mimic personalized tutoring in a computer-based environment and are a low-cost alternative to human tutors.
Over the past two decades, many ITS have been successfully deployed to enhance teaching and improve students' learning experience in a number of domains~\cite{AbuEl,Agha,Nakhal,Rekhawi,budenbender2002,goguadze2005interactivity,leelawong2008designing,melis2004,passier2006,Qwaider}, not only providing feedback and assistance but also addressing individual student characteristics~\cite{Graesser} and cognitive processes~\cite{Wu}.
Many ITS consider the development of a personalized curriculum and personalized feedback~\cite{Aldahdooh,Nakhal,albacete2019impact,chi2011instructional,Lin,munshi2019personalization,rus2014macro,rus2014deeptutor}, with dialogue-based ITS being some of the most effective tools for learning~\cite{ahn2018adaptive,graesser2005autotutor,graesser2001intelligent,nye2014autotutor,ventura2018preliminary}, as they simulate a familiar learning environment of student--tutor interaction, thus helping to improve student motivation. The main bottleneck is the ability of ITS to address the multitude of possible scenarios in such interactions, and this is where methods of automated, data-driven feedback generation are of critical importance.

Our paper has two major contributions.
Firstly, we describe how state-of-the-art machine learning (ML) and natural language processing (NLP) techniques can be used to generate automated, data-driven \textit{personalized hints and explanations}, \textit{Wikipedia-based explanations}, and \textit{mathematical hints}.
Feedback generated this way takes the individual needs of students into account, does not require expert intervention or hand-crafted rules, and is easily scalable and transferable across domains.
Secondly, we demonstrate that the personalized feedback leads to substantially improved student learning gains and improved subjective feedback evaluation in practice. To support our claims, we utilize our feedback models in {\tt Korbit}, a large-scale dialogue-based ITS.





\section{{\tt Korbit} Learning Platform}

\begin{figure}[t]
\includegraphics[width=\textwidth]{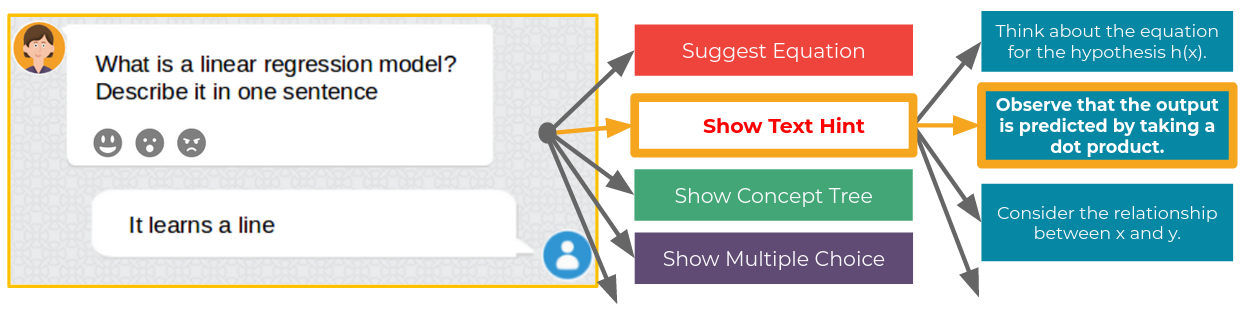}
\caption{An example illustrating how the {\tt Korbit} ITS {\tt inner-loop} system selects the pedagogical intervention. The student gives an incorrect solution and receives a text hint.} \label{intervention_flow}
\end{figure}

{\tt Korbit} is a large-scale, open-domain, mixed-interface, dialogue-based ITS, which uses ML, NLP and reinforcement learning to provide interactive, personalized learning online.
Currently, the platform has thousands of students enrolled and is capable of teaching topics related to data science, machine learning, and artificial intelligence.

Students enroll based on courses or skills they would like to study.
Once a student has enrolled, {\tt Korbit} tutors them by alternating between short lecture videos and interactive problem-solving.
During the problem-solving sessions, the student may attempt to solve an exercise, ask for help, or even skip it.
If the student attempts to solve the exercise, their solution attempt is compared against the expectation (i.e.\@ reference solution) using an NLP model.
If their solution is classified as incorrect, the {\tt inner-loop} system (see Fig. \ref{intervention_flow}) will activate and respond with one of a dozen different pedagogical interventions, which include hints, mathematical hints, elaborations, explanations, concept tree diagrams, and multiple choice quiz answers.
The pedagogical intervention is chosen by an ensemble of machine learning models from the student's zone of proximal development (ZPD)~\cite{cazden1979peekaboo} based on their student profile and last solution attempt.


\section{Automatically Generated Personalized Feedback}

In this paper, we present experiments on the {\tt Korbit} learning platform with actual students. These experiments involve varying the text hints and explanations based on how they were generated and how they were adapted to each unique student.

\vspace{-1.5em}
\subsubsection{Personalized Hints and Explanations} are generated using NLP techniques applied by a 3-step algorithm to all expectations (i.e.\@ reference solutions) in our database: (1) keywords, including nouns and noun phrases, are identified within the question (e.g. {\em overfitting} and {\em underfitting} in Table \ref{tab:text_hint_generation}); (2) appropriate sentence span that does not include keywords is identified in a reference solution using state-of-the-art dependency parsing with {\tt spaCy}\footnote{\url{https://spacy.io}} (e.g., {\em A model is underfitting} is filtered out, while {\em it has a high bias} is considered as a candidate for a hint); and (3) a grammatically correct hint is generated using discourse-based modifications (e.g., {\em Think about the case}) and the partial hint from step (2) (e.g., {\em when it has a high bias}).

\vspace{-1.5em}
\begin{table}
\centering
\caption{Hint generation. Keywords are marked with boxes}
\label{tab:text_hint_generation}
\begin{tabular}{lll}
{\bf Question} &  {\bf Expectation} & {\bf Generated hint} \\
\hline
What is the \fbox{difference} between \hspace{0.1cm} & A model is \fbox{underfitting} \hspace{0.10cm} & \underline{Think about the case} \\
\fbox{overfitting} and \fbox{underfitting}? &  when it has a high bias. &  when it has a high bias.\\ 
\end{tabular}
\end{table}
\vspace{-1.5em}



Next, hints are ranked according to their linguistic quality as well as the past student--system interactions. We employ a Random Forest classifier using two broad sets of features: (1) {\em Linguistic quality features} assess the quality of the hint from the linguistic perspective only (e.g., considering length of the hint/explanation, keyword and topic overlap between the hint/explanation and the question, etc.), and are used by the {\bf {\em baseline model}} only. (2) {\em Performance-based features} additionally take into account past student interaction with the system. Among them, the {\bf {\em shallow personalization model}} includes features related to the number of attempted questions, proportion of correct and incorrect answers, etc., and the {\bf {\em deep personalization model}} additionally includes linguistic features pertaining to up to $4$ previous student--system interaction turns. The three types of feedback models are trained and evaluated on a collection of $450$ previously recorded student--system interactions.

\vspace{-1.5em}
\subsubsection{Wikipedia-Based Explanations} provide alternative ways of helping students to understand and remember concepts.
We generate such explanations using another multi-stage pipeline: first, we use a 2 GB dataset on ``Machine learning'' crawled from Wikipedia and extract all relevant domain keywords from the reference questions and solutions using {\tt spaCy}. Next, we use the first sentence in each article as an {\em extracted Wikipedia-based explanation} and the rest of the article to {\em generate candidate explanations}.
A Decision Tree classifier is trained on a dataset of positive and negative examples to evaluate the quality of a Wikipedia-based explanation using a number of linguistically-motivated features.
This model is then applied to identify the most appropriate Wikipedia-based explanations among the generated ones.

\vspace{-1.5em}
\subsubsection{Mathematical Hints} are either provided by {\tt Korbit} in the form of suggested equations with gapped mathematical terms for the student to fill in, or in the form of a hint on what the student needs to change if they input an incorrect equation. Math equations are particularly challenging because equivalent expressions can have different representations: for example, $y$ in $y(x+5)$ could be a function or a term multiplied by $x+5$. To evaluate student equations, we first convert their {\LaTeX} string into multiple parse trees, where each tree represents a possible interpretation, and then use a classifier to select the most likely parse tree and compare it to the expectation. Our generated feedback is fully automated, which differentiates {\tt Korbit} from other math-oriented ITS, where feedback is generated by hand-crafted test cases~\cite{budenbender2002,hennecke1999online}.

\section{Experimental Results and Analysis}

Our preliminary experiments with the {\em baseline}, {\em shallow} and {\em deep personalization} models run on the historical data using $50$-fold cross-validation strongly suggested that {\em deep personalization} model selects the most appropriate personalized feedback. 
To support our claims, we ran experiments involving $796$ annotated student--system interactions, collected from $183$ students enrolled for free and studying the machine learning course on the {\tt Korbit} platform between January and February, 2020. 
First, a hint or explanation was selected at uniform random from one of the personalized feedback models when a student gives an incorrect solution. 
Afterwards, the student learning gain was measured as the proportion of instances where a student provided a correct solution after receiving a personalized hint or explanation.
Since it's possible for the ITS to provide several pedagogical interventions for a given exercise, we separate the learning gains observed for all students from those for students who received a personalized hint or explanation before their second attempt at the exercise.
Table \ref{table:personalized_hints_and_explanations_learning_gains} presents the results, showing that the {\em deep personalization model} leads to the highest student learning gains at $48.53\%$ followed by the {\em shallow personalization model} at $46.51\%$ and the {\em baseline model} at $39.47\%$ for all attempts.
The difference between the learning gains of the {\em deep personalization model} and {\em baseline model} for the students before their second attempt is statistically significant at 95\% confidence level based on a z-test (p=0.03005).
These results support the hypothesis that automatically generated personalized hints and explanations lead to substantial improvements in student learning gains.

\vspace{-1.5em}
\begin{table}[ht]
\centering
\caption{Student learning gains for personalized hints and explanations with 95\% confidence intervals (C.\@I.\@). 
}
\label{table:personalized_hints_and_explanations_learning_gains}
\begin{tabular}{lcccccc}
& \multicolumn{2}{c}{All Attempts} & \multicolumn{2}{c}{Before Second Attempt} \\
Model & Mean & \ 95\% C.\@I.\@ & Mean & 95\% C.\@I.\@ \\
\hline
Baseline \scriptsize{(No Personalization)} & $39.47\%$ & $[24.04\%, 56.61\%]$ & $37.93\%$ & $[20.69\%, 57.74\%]$ \\
Shallow Personalization & $46.51\%$ & $[31.18\%, 62.34\%]$ & $51.43\%$ & $[33.99\%, 68.62\%]$ \\
Deep Personalization & $\mathbf{48.53\%}$ & $\mathbf{[36.22\%, 60.97\%]}$ & $\mathbf{60.47\%}$ & $\mathbf{[44.41\%, 75.02\%]}$ \\ 
\hline \hspace{0.25cm}
\end{tabular}
\end{table}
\vspace{-2.5em}

Experiments on the {\tt Korbit} platform confirm that extracted and generated {\em Wikipedia-based explanations} lead to comparable student learning gains. Students rated either or both types of explanations as helpful $83.33\%$ of the time. This shows that automatically-generated Wikipedia-based explanations can be included in the set of interventions used to personalize the feedback.
Moreover, two domain experts independently analyzed a set of $86$ student--system interactions with {\tt Korbit}, where the student's solution attempt contained an incorrect mathematical equation. The results showed that over $90\%$ of the {\em mathematical hints} would be considered either ``very useful" or ``somewhat useful".

In conclusion, our experiments strongly support the hypothesis that the personalized hints and explanations, as well as Wikipedia-based explanations, help to improve student learning outcomes significantly. Preliminary results also indicate that the mathematical hints are useful. Future work should investigate how and what types of Wikipedia-based explanations and mathematical hints may improve student learning outcomes, as well as their interplay with student learning profiles and knowledge gaps.

\bibliographystyle{splncs04}
\bibliography{mybibliography}

\end{document}